\pdfoutput=1

\documentclass[11pt]{article}

\usepackage[]{acl}
\usepackage{pifont}
\newcommand{\xmark}{\ding{55}}%
\usepackage{times}
\usepackage[hang]{footmisc}
\usepackage{latexsym}
\usepackage{graphicx}
\usepackage{dblfloatfix}
\usepackage{subcaption}
\usepackage{tablefootnote}
\usepackage{multicol}
\usepackage{multirow}
\usepackage{adjustbox}
\usepackage{comment}
\usepackage{placeins}
\usepackage{caption}
\usepackage{subcaption}
\usepackage{mathtools}
\usepackage{amsmath}
\usepackage{algorithm}
\usepackage{soul}
\usepackage{algpseudocode}
\algblock{Input}{EndInput}
\algnotext{EndInput}
\algblock{Parameters}{EndParameter}
\algnotext{EndParameter}
\newcommand{\Desc}[2]{\State \makebox[2em][l]{#1}#2}
\usepackage[T1]{fontenc}

\usepackage[utf8]{inputenc}
\newcommand{\pg}[1]{\textcolor{pink}{[\bf\small PG: #1]}}
\newcommand{\tb}[1]{\textcolor{brown}{[#1]}}

\newcommand{\noteng}[1]{\textcolor{red}{[\bf\small NG: #1]}}
\usepackage{microtype}

\title{A Framework to Generate High-Quality Datapoints for Multiple Novel Intent Detection}

\author{$^1$Ankan Mullick\thanks{ Equal contribution} \qquad $^2$Sukannya Purkayastha\footnotemark[1] \hspace{0.01cm} \thanks{Work done while the author was a student at IIT Kharagpur} \qquad $^1$Pawan Goyal \qquad $^{1,3}$Niloy Ganguly\\ $^1$Department of Computer Science and Engineering, IIT Kharagpur \\ $^2$TCS Research, India\\  $^3$Leibniz University of Hannover, Germany \\\texttt{ankanm@kgpian.iitkgp.ac.in},       \texttt{sukannya.purkayastha@tcs.com}\\\texttt{\{pawang, niloy\}@cse.iitkgp.ac.in}}

\begin{document}
\maketitle
\begin{abstract}
Systems like Voice-command based conversational agents 
are characterized by a pre-defined set of skills or intents to perform user specified tasks.  In the course of time, newer intents may emerge requiring retraining.  
However, the newer intents may not be explicitly announced and need to be inferred dynamically. 
Thus, there are two important tasks at hand (a). identifying emerging new intents, (b). annotating data of the new intents so that the underlying classifier can be retrained efficiently. The tasks become specially challenging when a large number of new intents emerge simultaneously and there is a limited budget of manual annotation.  
In this paper, we propose \textbf{MNID} (\textbf{M}ultiple \textbf{N}ovel \textbf{I}ntent \textbf{D}etection) which is a cluster based  framework to detect multiple novel intents with budgeted human annotation cost. Empirical results on various benchmark datasets (of different sizes) demonstrate that MNID, by intelligently using the budget for annotation, outperforms the baseline methods 
in terms of accuracy and F1-score. 


\end{abstract}

\section{Introduction}

The conversational agents such as Amazon Alexa, Apple Siri  are characterised by the skill of understanding {\em intents} which help them  to efficiently handle  a user's query. For example, the query \textbf{`Will it be colder in Ohio'} requires getting the weather updates for the city `Ohio' and would be associated to the intent \textit{GetWeather}. The agents are trained with a pre-defined set of intents such as \{\textit{GetWeather}, \textit{RateBook}, \textit{BookRestaurant}\} so as to perform the goal-oriented user tasks. 
But with the passage of time, a user may be interested in performing newer tasks adding hitherto unknown intents. For example, \textbf{`Play some music from 1954'} would be associated to the intent \textit{PlayMusic} that may not be a part of the set of pre-defined intents.  



Emergence of novel intent detection has been periodically checked by different models in the last decade.  There are works on   incremental learning in dynamic environment for evolving new classes \cite{zhou2002hybrid,kuzborskij2013n,scheirer2012toward}. 
There are also several  approaches 
\cite{sun2016online,masud2010classification,haque2016sand,wang2020active,mu2017streaming,mu2017classification}
to  detect new classes in the form of outlier detection but they do not distinguish among multiple new class labels so  are not effective in novel multi-class detection. 
{\newcite{xia2018zero,siddique2021generalized} detect user intents using zero-shot generalized intent detection framework. However, they assume that the unseen intent class LABELS are already known, while in our case neither the number of unseen intent classes, nor the corresponding class labels are known.} 
The other line of works   \cite{xia2021incremental,halder2020task} supply the system with new intents, albeit with a limited amount of tagged data per class  and then have an efficient algorithm to 
incrementally learn new classes. These models work on the  assumption  that some instances of these new classes would be provided for model building. 
However, in a realistic setting,  the system  may not have any knowledge of the number and types of new intents appearing, it may at most understand that some new out-of-domain samples are generated.  So, the problem statement is to probe the incoming data wisely and \textbf{use minimum human intervention to identify all types of novel intents emerging and intelligently tag a limited set of data covering all discovered intents}, which can be be fed into a model for retraining.


More concretely 
the system is at first trained with an initial set of known intents; side-by-side an out-of-distribution (OOD) detector classifier is also trained to identify datapoints which do not fit the known intents. When substantial amount of such points are detected, the task is to (a) identify whether the points are originating from introduction of a single novel intent or multiple and (b) choose (a limited number of) samples to annotate so that the classifier can be retrained efficiently.

In order to determine the number of novel intents present in the OOD data, we undertake a clustering based approach with the idea that each cluster would represent a novel intent. By increasing the cluster number progressively, we can make a highly accurate estimate of the number of novel intents. If sample points of an intent mainly correspond to a well formed cluster, the implication is that without much probing we can shortlist enough training samples  (through silver tagging) for that class. On the other hand, if the sample points of an intent tend to interwine with other intent points in the feature space, these can be considered as uncertain points and require human intervention for tagging (gold tagging). 
With this intuition in place, we design a mix of silver and gold tagging to produce high-quality training samples which can be used to retrain the classifier.  

Our proposed framework of Multiple Novel Intent Detection (\textbf{MNID}) is compared with competitive baselines and evaluated across several standard public datasets in NLU domain where it performs substantially better. We use datasets with different number of intent classes. SNIPS \cite{coucke2018snips} and ATIS {\cite{tur2010left}} are smaller datasets, consisting of less number intent classes - 7 and 21, respectively. HWU~\cite{liu2019benchmarking}, BANKING \cite{casanueva2020efficient} and CLINC \cite{larson2019evaluation} 
consist of large number of intent classes - 64, 77 and 150, respectively.


The paper is organized as follows. We discuss the Problem Setting and solution overview in Section \ref{sec:problem}. Our algorithmic framework is described in Section \ref{sec:solution}. We present the datasets with experimental statistics and data pre-processing in Section \ref{sec:dataset}. In Section \ref{sec:experiment}, we discuss the experimental design and baselines. 
Detail evaluation results with different algorithmic variations are in Section \ref{results}. We conclude
with a summary in Section \ref{sec:conclusion}\footnote{Codes are in -  \url{https://github.com/sukannyapurkayastha/MNID}}. 

\begin{algorithm}[!thb]
\caption{Multiple Novel Intent Detection (\textbf{MNID})} 
\label{algo:mnid}
\begin{algorithmic}[1]
\Input
  \Desc{$D_{init}$}{\hspace{2mm}Initial Labelled Data}
   \Desc{$\mathcal{T}$}{\hspace{2mm}Blind Test Data For Evaluation}
   \Desc{$B$}{\hspace{2mm}Total Annotation Budget}
  \EndInput
\Parameters
    \Desc{$D$}{\hspace{2mm} Total Data points}
 \Desc{$\mathcal{L}$}{$\gets D_{init}$}
\Desc{$\mathcal{OS}$}{$\gets$ OODD($D$, $D_{init}$)}
\EndParameter
\Procedure{Multiple Novel Intent Detection}{}
\State $\mathcal{L}$, $N_{new}$, $\mathcal{CL} \gets $ NCD($\mathcal{OS}, \mathcal{L}$)
\If{$\mathcal{L}<B$}
\State $\mathcal{L}$, $\mathcal{G_{CL}}$, $\mathcal{B_{CL}}\gets$ CBQA($\mathcal{L}$, $\mathcal{CL}$)
\EndIf
\State Train Model $\mathcal{M}$ on $\mathcal{L}$, predict on the remaining points in the clusters to get the confidence score (CS) of each data point and store in $All_{CS}$.

\If{$\mathcal{L}<B$}
\State $\mathcal{L} \gets$ PPAS($\mathcal{L}$, $All_{CS}$, $\mathcal{G_{CL}}$, $\mathcal{B_{CL}}$, $B$)
\EndIf
\State Train  model $\mathcal{M}$ on $\mathcal{L}$ and test on $\mathcal{T}$ to find out Accuracy, F1 for all classes. 
\EndProcedure
\end{algorithmic}
\end{algorithm}

\begin{algorithm}[!thb]
\caption{OOD Detection Algorithm \\ \textbf{OODD($|D|$, $|D_{init}|$)}}
\label{algo1:mnid-alc}
\begin{algorithmic}[1]
\State Train OOD-SDA on $D_{init}$ and predict on $(D-D_{init})$ to get OOD samples, $\mathcal{OS}$ 
\State Return $\mathcal{OS}$
\end{algorithmic}
\end{algorithm}

\begin{algorithm}[!thb]
\caption{Novel Class Detection \textbf{NCD($\mathcal{OS}$, $\mathcal{L}$)}}
\label{algo2:ncd}
\begin{algorithmic}[1]
\State Initial number of clusters, $K=1$.
\State Number of new classes, $N_{new}  = 0$. 
\While{$N_{new}\geq \lfloor K/2 \rfloor$}
\State Perform $K$-Means Clustering on $\mathcal{OS}$
\State Annotate $x$ ($\ge$ 2) points from each cluster. That results in discovering of $n'$ new classes 
\State Add $x*K$ point labels to $\mathcal{L}$
\State $N_{new} \gets N_{new} + n'$
\State $K \gets 2*K$
\EndWhile
\State $\mathcal{CL}$ = Store All $K$ Clusters
\State Return ($\mathcal{L}$,$N_{new}$, $\mathcal{CL}$)
\end{algorithmic}
\end{algorithm}

\begin{algorithm}[!thb]
\caption{Cluster Quality Based Annotation \textbf{CQBA($\mathcal{L}$, $\mathcal{CL}$)}}
\label{algo3:quality_cluster}
\begin{algorithmic}[1]
\State Take $p$ points from each of the clusters ($\mathcal{CL}$) and annotate to find Good Cluster ($\mathcal{G_{CL}}$) and Bad Cluster ($\mathcal{B_{CL}}$). 
\State Add annotated $p*|\mathcal{CL}|$ point labels to $\mathcal{L}$. 
\For{each Bad Cluster}
    \State Take $q$ more points from Bad cluster and annotate. 
    \State Add $q*|\mathcal{B_{CL}}|$ point labels to $\mathcal{L}$.
\EndFor
\State Return($\mathcal{L}$, $\mathcal{G_{CL}}$, $\mathcal{B_{CL}}$)
\end{algorithmic}
\end{algorithm}

\begin{algorithm}[!thb]
\caption{Post-Processing Annotation Strategy \textbf{PPAS($\mathcal{L}$, $All_{CS}$, $\mathcal{G_{CL}}$, $\mathcal{B_{CL}}$, $B$)}}
\label{algo4:annotation}
\begin{algorithmic}[1]
    \For{Each point with CS in $All_{CS}$} 
    \If{$CS \geq \mathcal{TH} $ and point in $\mathcal{G_{CL}}$ and average cosine similarity with already annotated points of that class $\geq \tau$}
    \State $L_{s} \gets$ Silver Annotation Strategy 
    \EndIf
    \EndFor
    \While{$|\mathcal{L}|<B$}
    \State Select datapoint with least CS 
        \If{$\mathcal{B_{CL}}$ exists}
        \State From $\mathcal{B_{CL}}$ in Round-Robin way
        \Else
        \State{From $\mathcal{G_{CL}}$ in Round-Robin way}
        \EndIf
        \State $L_{g} \gets$ Gold Annotation Strategy
        \State $\mathcal{L}$ $\gets$ $\mathcal{L}$ + $L_{g}$
    \EndWhile
\State $\mathcal{L} \gets$ $\mathcal{L}$ $\cup$ $L_s$ $\cup$ $L_g$ \\
Return ($\mathcal{L}$)
\end{algorithmic}
\end{algorithm}

\section{Problem Setting and Solution Overview}
\label{sec:problem}
{\bf Problem Setting:} To formally describe the problem setting, let there be a dataset $W$ containing overall $N$ classes. However, the value of $N$ is not known apriori.  Let $T\in W$ be the test set and $W - T = D$ be the rest of the dataset, out of which $|D_{init}|~(<< |D|)$ labelled data of $N_{init}~( < N)$ classes is initially provided, while the rest of the data is unlabelled. The task is to design an algorithm to (a). detect all the remaining $N - N_{init}$ classes and (b). 
spent  a limited budget ($B$ - $|D_{init}|$) to annotate high fidelitous new datapoints, so that the classifier can achieve high accuracy when retraining. 

\noindent{\bf Solution Overview:} 
The solution steps are as follows: (a) Identify the OOD (out of distribution) datapoints which do not belong to the initial $N_{init}$ classes. This can be considered as a preprocessing step. (b) Use a part of the allotted budget to annotate a portion of these OOD datapoints. These points (for annotation) are selected by repeatedly running a clustering algorithm with increasing number of clusters as input, and choosing cluster 
cluster centre points to identify the unknown classes. {\bf Rationale:} The intuition/expectation is that each cluster hosts a separate intent, hence annotating the cluster 
centres would lead to discovery of maximum number of novel intents.  (c) Further identify the classes which are well clustered in feature space and which are not. Use another portion of the budget to increase the annotations of not-so well formed clusters and then build up a classifier with all the classes. {\bf Rationale:} If a cluster is well-formed, most likely it is hosting a single class, hence there is no need to annotate further points there, rather 
annotate more points in not-so-well-formed clusters. 
(d) Use the classifier to classify points from the clusters.  Identify low-confidence points from the bad clusters and annotate them. High-confidence points from good clusters are  silver annotated. 
{\bf Rationale:} The low-confidence points in the bad clusters are the most uncertain points, hence  annotating them helps in increasing classifier accuracy. Similarly high-confident points in the good clusters almost surely will belong to that particular cluster, hence silver annotation is pursued.  
(e). Retrain the classifier. 

\begin{figure}
    \centering
    \includegraphics[width=7cm, height = 9 cm]{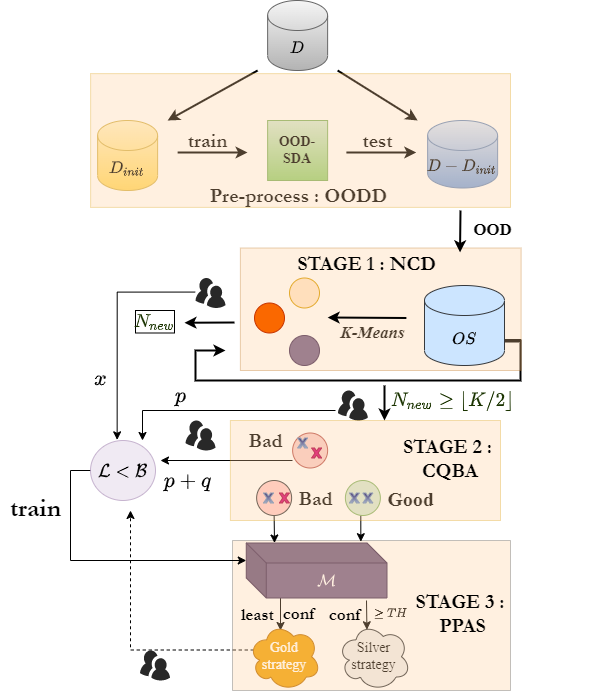}
    \caption{End-to-end architecture of MNID: Multiple \\Novel Intent Detection 
    }
    \label{fig:mnid-archi}
    \vspace{-2mm}
\end{figure}

The overall MNID framework with different algorithmic modules is shown in Fig. \ref{fig:mnid-archi}.

\section{MNID: Solution Detail}
\label{sec:solution}
The proposed framework for Multiple Novel Intent Detection (MNID) is explained through Algorithm~\ref{algo:mnid}. As highlighted in the overview, the algorithm consists of data pre-processing step, followed by three stages, each of them are discussed below. 
The total budget of (gold) annotation is $B$. Besides the system can undertake unlimited silver annotation. The advantage of silver strategy is that it is free as no human probing is required. However, it is also likely to bring in noise if used indiscriminately.   


\noindent \textbf{Pre-process: OOD Detection (OODD)}: For the dataset ($D$), this module (Algorithm \ref{algo1:mnid-alc}) takes the initial labelled data ($D_{init}$) as input and predicts the Out-of-Domain (OOD) samples on the remaining data, ($D-D_{init}$). We call the set of OOD samples predicted as ${\mathcal {OS}}$. This is a part of the data pre-processing.

\noindent \textbf{Stage 1. Novel Class detection (NCD)}: In this sub-module (Algorithm \ref{algo2:ncd}), we aim at finding all the new classes, $N_{new}.$ On the OOD samples (${\mathcal {OS}}$), obtained in the previous sub-module (Algorithm~\ref{algo1:mnid-alc}), we do clustering using K-Means. We start the algorithm with $K = 1$ and number of new classes, ${N}_{new}$ = 0. We perform - (i) K-Means clustering. (ii) Annotate $x$ points from each cluster, add those points to $\mathcal{L}$ and identify $n'$ new classes. (iii) Increase new class count (${N}_{new}$+ $n'$). (iv) Double the number of cluster count (we compare ${N}_{new}$ with $K/2$). We execute the above steps until  cluster count exceeds the new intent count. The algorithm returns current annotations ($\mathcal{L}$), newly discovered class count (${N}_{new}$) and newly formed clusters ($\mathcal{CL}$). The budget spent in this step is $B_1$. 

\noindent \textbf{Stage 2. Cluster Quality Based Annotation (CQBA)}: In this step (Algorithm \ref{algo3:quality_cluster}), we evaluate the quality of each of the clusters obtained by the previous algorithm. We annotate $p$ points from each of these clusters and if all the $p$ points belong to the same class, we term it as a {\sl good cluster} or else a {\sl bad cluster}. An example of a {\sl bad cluster} in BANKING would be the one containing data points from multiple classes, which may have high similarity, such as: \textit{declined\_cash\_withdrawal} and \textit{pending\_cash\_withdrawal}.
 For the {\sl bad clusters}, we annotate $q$ more points. All these annotated points are then added to the labelled data, $\mathcal{L}$. The budget spent in this step is $B_2$. Hence the remaining budget $B$ - ($B_1$ + $B_2$) is used in the next step.   \\
\noindent \textbf{Stage 3. Post Processing Annotation Strategy (PPAS)}: In this step (Algorithm \ref{algo4:annotation}), we add more data to the labelled set, $\mathcal{L}$, through gold annotation ({\sl gold strategy}), as well as silver-annotated data ({\sl silver strategy}). To select these data points, we first train a classifier ($\mathcal{M}$) with the labelled set, $\mathcal{L}$ as obtained in the last step (CBQA), and consider the clusters $\mathcal{CL}$. We predict on the remaining points of the clusters to get the confidence of the datapoints. We perform silver strategy based on confidence score (CS) and gold strategy in round-robin way 
to operate on each cluster one after another.\\
\textbf{Gold Strategy}: Least confident data-points are annotated from the bad clusters (if present) 
or else from the good clusters. Gold strategy is performed in a round-robin way 
to retrieve data points with the least score for each cluster
until our budget exhausts.\\
\textbf{Silver strategy}: 
If the confidence score (CS) of a datapoint from a cluster is greater than a predefined threshold ($\mathcal{TH}$), we measure the average cosine similarity of points annotated within that cluster with this point. If similarity is above a predefined threshold ($\mathcal{\tau}$), we label this point with class label of that cluster. 
The predefined threshold ($\mathcal{\tau}$) is required to choose good samples selectively instead of choosing all the points. Silver strategy does not require human intervention therefore there is no extra addition to the annotation cost, but the multiple conditions are checked to prevent noise in the training set.  
\noindent\textbf{Final Step:} We again train the neural model $\mathcal{M}$ on $\mathcal{L}$ and test on $\mathcal{T}$ to find out Accuracy and F1.

\section{Dataset and Pre-Processing}
\label{sec:dataset}

We perform our experiments on a variety of datasets, which are widely used as benchmarks for Natural Language Understanding tasks. The datasets are 
SNIPS \cite{coucke2018snips}, ATIS \cite{tur2010left}, 
HWU~\cite{liu2019benchmarking}, BANKING \cite{casanueva2020efficient} and CLINC \cite{larson2019evaluation}. SNIPS (7) and ATIS (21) are smaller datasets consisting of less number of intents (in bracket) where HWU (64), BANKING (77) and CLINC (150) are larger datasets with many intents. ATIS is the most imbalanced, skewed dataset. In BANKING data - several intents are highly similar among themselves. 
The detailed statistics of these datasets including our experimental framework are shown in Table \ref{tab:dataset}. \ul{Since the datasets are already  fully labelled, annotation essentially means utilizing the already available labels. Hence, we do not have to deal with usual issues of annotation accuracy, inter-annotator agreement, etc. }


\begin{table}[]
\centering
\captionsetup{justification=centering,margin=-0.25mm}
\begin{adjustbox}{width=\linewidth}
\begin{tabular}{|c|c|c|c|ccc|c|}
\hline
\multirow{2}{*}{\textbf{Dataset ($W$)}} & \multirow{2}{*}{\textbf{\begin{tabular}[c]{@{}c@{}}\# Intent \\ Class ($|N|$)\end{tabular}}} & \multirow{2}{*}{\textbf{\begin{tabular}[c]{@{}c@{}}Dataset \\ Size ($|W|$)\end{tabular}}} 
& \multirow{2}{*}{\textbf{\begin{tabular}[c]{@{}c@{}}\#Labelled  \\ |$D_{init}$|\end{tabular}}} & \multicolumn{3}{c|}{\textbf{\#Unlab ($|D|$ - |$D_{init}$|)}} & \multirow{2}{*}{\textbf{\begin{tabular}[c]{@{}c@{}}\#Test \\ ($|T|$))\end{tabular}}} \\ \cline{5-7}

                                  &                                                                                      &                                                                                                               &                                 & \multicolumn{1}{c|}{\textbf{\#IND}} & \multicolumn{1}{c|}{\textbf{\#OOD}} & \textbf{\#Total} &                                  \\ \hline
SNIPS                              & 7 (5+2)                                                                            & 14484                                                                            
& 50                             & \multicolumn{1}{c|}{8601}           & \multicolumn{1}{c|}{3449}            & 12050             & 2384                             \\ \hline
ATIS                              & 21 (13+8)                                                                            & 5871                                                                            
& 130                             & \multicolumn{1}{c|}{3155}           & \multicolumn{1}{c|}{1586}            & 4741             & 1000                             \\ \hline
HWU*                           & 64 (10+54)                                                                         & 11036                                                                           
& 100                             & \multicolumn{1}{c|}{1408}            & \multicolumn{1}{c|}{8452}           & 9860            & 1076                             \\ \hline
BANKING*                          & 77 (10+67)                                                                           & 13083                                                                          
& 100                             & \multicolumn{1}{c|}{1026}           & \multicolumn{1}{c|}{8877}           & 9903             & 3080                             \\ \hline
CLINC*                           & 150 (10+140)                                                                         & 22500                                                                           
& 100                             & \multicolumn{1}{c|}{1100}            & \multicolumn{1}{c|}{16800}           & 17900            & 4500                             \\ \hline
\end{tabular}
\end{adjustbox}
\caption{Statistics based on our split for five Datasets. \\ * represents pre-defined train-test splits. In \# Intent Class, (- + -) represents (known + unknown) intents} 
\vspace{-5mm}
\label{tab:dataset}
\end{table}

\if{0}
\noindent \textbf{SNIPS}~\cite{coucke2018snips}: This dataset is collected from the Snips personal voice assistant. It has a total of 14,484 sentences with 7 intent types. 
\\
\noindent \textbf{ATIS}~\cite{tur2010left}: ATIS consists of recordings from automated airline travel inquiry systems. We consider the same data as in \cite{goo-etal-2018-slot} for our experiments similar to \cite{chen2019bert}. It has total 5,871 sentences and 21 intent types.


\noindent \textbf{BANKING}~\cite{casanueva2020efficient}: All intents in this dataset are taken from the `Banking' domain. It has total 77 intents across 13,083 sentences.

\noindent \textbf{CLINC}~\cite{larson2019evaluation}: CLINC is a multi-domain dataset where there are 150 intents which span across 10 domains. The dataset contains queries labelled with these intents and some of them are labelled as `OOS', we however do not consider these OOS queries for our setting. It has total 22500 sentences with 150 intent types.

\noindent \textbf{HWU}~\cite{liu2019benchmarking}: 
It contains queries spanning multiple domains,
unlike a single domain SDS which only understands and responds to the user in a specific domain.
It consists of 11,036 examples for 64 intents in 21 domains. 
\fi

\subsection*{Data Pre-Processing}
\label{pre-process}

\begin{table}[!hbt]
\vspace{-3mm}
\centering
\captionsetup{justification=centering,margin=-0.1mm}
\begin{adjustbox}{width=\linewidth}
\begin{tabular}{|c|cc|cc|cc|cc|}
\hline
\multirow{2}{*}{\textbf{Dataset}} & \multicolumn{2}{c|}{\textbf{DOC}}               & \multicolumn{2}{c|}{\textbf{MSP}}               & \multicolumn{2}{c|}{\textbf{LMCL}}              & \multicolumn{2}{c|}{\textbf{FS-OOD}}            \\ \cline{2-9} 
                                  & \multicolumn{1}{c|}{\textbf{A}} & \textbf{F1} & \multicolumn{1}{c|}{\textbf{A}} & \textbf{F1} & \multicolumn{1}{c|}{\textbf{A}} & \textbf{F1} & \multicolumn{1}{c|}{\textbf{A}} & \textbf{F1} \\ \hline
SNIPS                           & \multicolumn{1}{c|}{\textbf{77.3}}             &       72.1      & \multicolumn{1}{c|}{78.2}             &    71.7               & \multicolumn{1}{c|}{74.7}             &     69.3    & \multicolumn{1}{c|}{76.8}         & \textbf{72.9}  \\ \hline
ATIS                              & \multicolumn{1}{c|}{55.8}        & 47.2           & \multicolumn{1}{c|}{56.1}             &    44.7         & \multicolumn{1}{c|}{54.5}             &     40.6        & \multicolumn{1}{c|}{\textbf{74.9}}        & \textbf{68.6}       \\ \hline
HWU                            & \multicolumn{1}{c|}{61.4}             &    57.2       & \multicolumn{1}{c|}{59.9}             &      29.9      & \multicolumn{1}{c|}{53.1}             &    44.3        & \multicolumn{1}{c|}{\textbf{68.2}}        &    \textbf{64.1}    \\ \hline
BANKING                           & \multicolumn{1}{c|}{56.3}             &    20.4         & \multicolumn{1}{c|}{52.5}             &    20.2         & \multicolumn{1}{c|}{52.9}             &    51.3         & \multicolumn{1}{c|}{\textbf{73.7}}        & \textbf{64.1}       \\ \hline
CLINC                            & \multicolumn{1}{c|}{54.8}             &    18.7         & \multicolumn{1}{c|}{53.4}             &      20.5       & \multicolumn{1}{c|}{54.1}             &    59.9         & \multicolumn{1}{c|}{\textbf{77.7}}        & \textbf{65.7}       \\ \hline

\end{tabular}
\end{adjustbox}
\caption{Accuracy (A) and F1-Score in (\%) of various OODD algorithms to detect OOD points from different datasets. \textbf{Bold} denotes the best for each dataset.}
\vspace{-3mm}
\label{tab:tab_iclr}
\end{table}

In the pre-processing step, we filter the out-of-domain samples. We consider four algorithms for detecting out-of-domain samples. i) Softmax Prediction Probability (MSP) \cite{hendrycks2018baseline} predicts out-of-domain samples based on a threshold on the softmax prediction scores. ii) Deep Open Classification (DOC) \cite{shu2017doc} method builds a multi-class classifier with one vs rest layer of sigmoids. iii) Large Margin Cosine Loss~(LMCL) \cite{lin2019deep} trains a network with margin loss 
and predictions are then fed into an algorithm called Local Outlier Factor~(LOF) for outlier detection. 
iv) Few-shot OOD (FS-OOD)~\cite{tan2019out} uses a ProtoTypical Network to detect OOD examples and classifying in-domain examples with few-shot examples from the in-domain class. 
We fine-tune BERT embeddings using all these out of domain sample detection algorithms. We use bert-base-uncased for these methods. We set the threshold for MSP as 0.5 as in \newcite{lin2019deep}, \newcite{xu2020deep}. The results of all these algorithms are shown in Table \ref{tab:tab_iclr}. FS-OOD~\cite{tan2019out} provides us the best accuracy and F1 for detecting OOD samples ($\mathcal{OS}$). Only DOC performs better in case of SNIPS but overall FS-OOD outperforms other approaches so we use FS-OOD produced out-of-sample data.\footnote{FS-OOD: \url{https://github.com/SLAD-ml/few-shot-ood} and other OOD models: \url{https://huggingface.co}} 

\section{Experimental Setup}
\label{sec:experiment}

The efficacy of the algorithm needs to be tested on two aspects. (a). The number of unknown intents identified.  (b). The accuracy achieved when the data is annotated by our algorithm, MNID. To test the accuracy, we use state-of-the-art several classification algorithms used for intent detection. 

\begin{table*}[]
\centering
\begin{adjustbox}{width=0.9\linewidth}
\begin{tabular}{|cc|c|c|c|c|}
\hline
\multicolumn{2}{|c|}{Method}                     & IFSTC ($Gl_{F}$, $Rn_{F}$, MNID) & PolyAI ($Gl_{F}$, $Rn_{F}$, MNID) & BERT ($Gl_{F}$, $Rn_{F}$, MNID)  & RoBERTa ($Gl_{F}$, $Rn_{F}$, MNID) \\ \hline
\multicolumn{1}{|c|}{\multirow{2}{*}{SNIPS}} & A & \textbf{85.4}, 78.1, 84.7 & 93.2, 85.7, \textbf{95.1} & 92.7, 91.6, \textbf{93.3} & 94.9, 92.3, \textbf{95.3} \\ \cline{2-6} 
\multicolumn{1}{|c|}{}                  & F1 & \textbf{84.2}, 79.4, \textbf{84.2} & 93.2, 84.3, \textbf{94.9} & 92.6, 91.9, \textbf{93.9} & \textbf{94.8}, 91.9, \textbf{94.8} \\ \hline
\multicolumn{1}{|c|}{\multirow{2}{*}{ATIS}} & A & 88.4, 70.1, \textbf{88.8} & 87.8, 71.8, \textbf{88.6} & 88.1, 70.2, \textbf{88.2} & 87.9, 70.8, \textbf{88.6} \\ \cline{2-6} 
\multicolumn{1}{|c|}{}                  & F1 & 87.3, 65.8, \textbf{87.8} & 84.3, 74.5, \textbf{87.0} & 86.3, 73.9, \textbf{86.9} & 84.6, 74.1, \textbf{85.1} \\ \hline
\multicolumn{1}{|c|}{\multirow{2}{*}{HWU}} & A & 78.2, 72.4, \textbf{79.7} & 83.8, 75.2, \textbf{83.8} & 82.6, 73.6, \textbf{82.7} & 82.5, 75.3, \textbf{83.7} \\ \cline{2-6} 
\multicolumn{1}{|c|}{}                  & F1 & 76.4, 71.4,
\textbf{78.4} & 83.7, 77.3, \textbf{84.2} & 81.7, 74.3, \textbf{82.4} & 81.3, 77.2, \textbf{82.4} \\ \hline
\multicolumn{1}{|c|}{\multirow{2}{*}{BANKING}} & A & 78.3, 72.8, \textbf{79.0} & 84.2, 79.0, \textbf{84.7} & 80.1, 75.5, \textbf{82.8} & 83.4, 77.0, \textbf{84.5} \\ \cline{2-6} 
\multicolumn{1}{|c|}{}                  & F1 & 77.7, 74.1, \textbf{80.0} & 83.1, 79.0, \textbf{84.4} & 80.0, 76.4, \textbf{83.7} & \textbf{83.9}, 78.7, 83.8  \\ \hline
\multicolumn{1}{|c|}{\multirow{2}{*}{CLINC}} & A & 88.7, 77.1, \textbf{88.9} & 92.1*, 83.2, \textbf{94.9} & 90.8, 77.6, \textbf{91.4} & 91.3, 84.5, \textbf{92.3} \\ \cline{2-6} 
\multicolumn{1}{|c|}{}                  &  F1& 85.7, 76.4, \textbf{88.3} & 93.5, 83.7, \textbf{95.2} & 90.7, 78.8, \textbf{91.0} & 91.7, 85.3, \textbf{92.8} \\ \hline
\end{tabular}
\end{adjustbox}
\caption{Overall Accuracy (A) and Macro F1 in (\%) across all datasets for different scenarios - ideal ($Gl_{F}$), random ($Rn_{F}$) and MNID (The best outcomes among three scenarios in \textbf{Bold}). *\newcite{casanueva2020efficient} report accuracy of 90.15 with OOS and 92.14 without OOS. 
} 
\label{tab:embeddings}
\end{table*}



\noindent \textbf{Different Neural Models:}
We explore different neural models to evaluate {\bf MNID} as discussed next:

\textbf{1. IFSTC~\cite{xia2021incremental}:} This finetunes a trained model on few shot data of new classes using an entailment and hybrid strategy. We use the hybrid strategy (best performing in their case).

  \textbf{2. PolyAI~\cite{casanueva2020efficient}:} It performs intent classification task based on dual sentence encoders - Universal Sentence Encoders (USE)~\cite{cer2018universal} and ConveRT. Since authors have taken down the ConveRT model, we apply USE only. \footnote{We use author's implementation of IFSTC (PyTorch) and re-implement PolyAI (Tensorflow)}

 Along with the above two, we also consider other standard models, \textbf{3. BERT (`bert-base-uncased')}~\cite{devlin2019bert} and \textbf{4. RoBERTa (`roberta-base-uncased')}~\cite{Liu2019RoBERTaAR} for evaluation on these datasets. We finetune these pre-trained language models for 15 epochs for the smaller datasets (SNIPS, ATIS) 50 epochs for the larger datasets 
 {(HWU, BANKING, CLINC)} and  with a learning rate of 2e-05 and Adam optimizer\footnote{We use \url{https://huggingface.co/}}. 
 { Early stopping was employed to stop training.} 
 For all methods, we provide the same number of gold annotated data obtained using our pipeline and report its performance.  


\noindent \textbf{Baselines}: We compare the performance of our method using two annotation techniques for choosing $B - \|D_{init}\|$ data points: 1) \textbf{\textit{$Gl_F$}}: This is the ideal scenario where we are given $F$ (=10) data points for each of the new classes - \textit{Gold\textsubscript{Few}}, abbreviated as `$Gl_F$'. 2) \textbf{\textit{$Rn_{F}$}} : 
Here, we randomly choose $B - \|D_{init}\|$ data points from the unlabelled data - \textit{Random\textsubscript{Few}}, abbreviated as `$Rn_{F}$'.




\noindent \textbf{Clustering Algorithms}: One of the building blocks of \textbf{MNID} is to cluster datapoints, so the efficacy of \textbf{MNID} depends on employing an efficient clustering algorithm. We do a detailed study by employing several unsupervised and semi-supervised clustering algorithm and choose the best. 

The unsupervised algorithms are: (i) \textbf{K-Means (KM)} \cite{macqueen1967some} (ii) \textbf{Agglomerative Clustering (AG)} \cite{gowda1978agglomerative} (iii) \textbf{Deep Clustering Network (DCN)} \cite{yang2017towards} and (iv) \textbf{Deep Embedded Clustering (DEC)} \cite{xie2016unsupervised} which uses the stacked auto-encoder based reconstruction loss.
The semi-supervised algorithms are: (i) \textbf{DeepAligned (DAL)} \cite{zhang2021discovering} which uses limited data for pre-training and cluster assignments as pseudo labels for cluster refinement. (ii) \textbf{DTC} \cite{han2019learning} develops on the DCN algorithm by scaling it to the transfer learning setting and can estimate the number of known classes in unlabelled data. It is however highly dependent on availability of labelled data (iii) \textbf{KCL} \cite{hsu2017learning} which transfers the knowledge to target dataset considering KL-divergence based distance loss (iv) \textbf{MCL} \cite{hsu2019multi} which uses meta-classification based likelihood criterion for pairwise similarity evaluation (v) \textbf{CDAC+} \cite{lin2020discovering} which uses prior data to refine the clustering process and KL-divergence based loss  \footnote{Code: \url{https://github.com/thuiar/TEXTOIR} }.
Other than KM and AG, all the other unsupervised methods along with some of the semi-supervised methods such as DTC and CDAC need the information of the ground truth number of clusters for training and we provide them so (it is an extra advantage for them). For semi-supervised methods such as KCL, MCL, DTC and DAL, we start with double the number of ground truth clusters and let the method determine the number of clusters.

\noindent \textbf{Hyper-parameters and Settings:}
For Post-Processing annotation strategy of MNID, we set the cosine similarity threshold, $\tau$ as 0.8 and the confidence threshold, $\mathcal{TH}$ as 0.5 \footnote{This combination of $\tau$ and $\mathcal{TH}$ provides the best results among different experimented results.}. For all datasets, we use a setting similar to $\kappa$-shot with $\kappa$ = 10. For $N$ intents, we define our total budget $B = \kappa \times N$. We use same budget for all our experiments. We experiment on NVIDIA Tesla K40m GPU with 12 GB RAM, 6 Gbps clock cycle and GDDR5 memory. All the methods took less than 8 GPU hours for training. 

\begin{figure*}[!b]
\centering
        \begin{subfigure}[b]{0.3\textwidth}
                \centering
                \includegraphics[width=.95\linewidth]{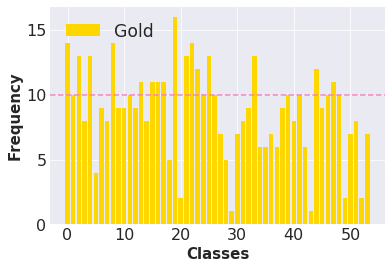}
                \caption{HWU}
                \label{fig:hwu}
        \end{subfigure}%
        \begin{subfigure}[b]{0.3\textwidth}
                \centering \includegraphics[width=.95\linewidth]{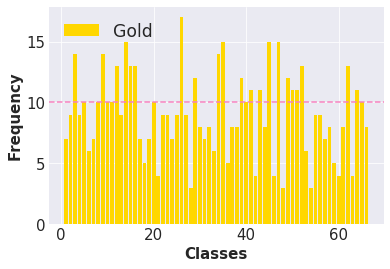}
                \caption{BANKING}
                \label{fig:bank}
        \end{subfigure}%
        \begin{subfigure}[b]{0.3\textwidth}
                \centering
                \includegraphics[width=.95\linewidth]{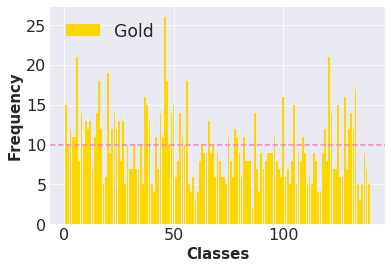}
                \caption{CLINC}
                \label{fig:clinc}
        \end{subfigure}%
        \caption{Count of gold annotated points for newly discovered classes} 
        \label{fig:class_dist}
\end{figure*}

\section{Experimental Results}
\label{results}

In this section, we discuss the experimental outcomes for MNID and competing baselines. We also show results of different clustering algorithms and variations of distinct components of MNID. 

\noindent \textbf{(A) Class Discovery:} MNID is very effective in identifying almost all new intents. For HWU, BANKING and CLINC,   54 out of 54 (100\%), 66 out of 67 (98.5\%) and 139 out of 140 (99.3\%) new intents from the unknown intent set were discovered, respectively. For SNIPS and ATIS, we could discover 2 out of 2 (100\%) and 7 out of 8 (87.5\%) new intent classes, respectively. Due to data skewness (ATIS) and high intent similarity (BANKING, CLINC) MNID misses one intent.
\label{results}
\noindent \textbf{(B) Performance of MNID:} Table \ref{tab:embeddings} shows the performance of different models - IFSTC, PolyAI, BERT and RoBERTa when trained with datasets provided by MNID. In order to maintain the fairness, MNID, $Rn_F$ and $Gl_F$ use the (overall) same number of gold-annotated data points. Besides MNID uses silver-annotated data points, while the others do not have any way of creating high quality silver annotated data. Each cell in the table consists of values from $Gl_{F}$, $Rn_{F}$ and MNID. As expected, $Rn_{F}$ performs the worst across all settings. However, except two scenarios, we observe that MNID consistently performs better than the $Gl_F$ dataset. For all these four different settings across five datasets, MNID improvements over $Gl_{F}$ predictions are statistically significant ($p < 0.05$) as per McNemar's Test. 
It is observed that our approach also works well on the highly imbalanced ATIS dataset in which some of the classes have less than 10 data points and highly similar BANKING dataset in which the intents are closely related  eg., `top-up-reverted' and `top-up-failed'.
This is because although $Gl_F$ chooses uniformly across all classes, MNID selectively labels datapoints having high uncertainty thus providing the classifier with the right ingredient to perform better. 
In IFSTC on SNIPS dataset, MNID underperforms as compared to $Gl_F$ but with a very small margin. This happens because in the case of SNIPS dataset,  the number of new classes is very less, hence $Gl_{F}$ can choose ideal candidates. 
The best performance of MNID as well as the two baselines is in the PolyAI setting when it is used with Universal Sentence Encoders (USE). 
Since PolyAI performs the best, all our subsequent results are provided on PolyAI (USE). 

\noindent \textbf{(C) Distribution of gold annotated points:} 
Fig \ref{fig:class_dist} shows the count of the gold annotated points ($Y-axis$) for new classes (class indices on $X-axis$). 
The dotted line is at the frequency of 10, corresponding to the average annotations per class. For 76.2\% (HWU), 81.5\% (BANKING) and 67.3\% (CLINC) classes in good clusters require `$\leq 10$' annotations. More than 10 annotations are needed for 65.4\% (HWU), 68.5\% (BANKING) and 54.5\% (CLINC) classes in bad clusters.  

\noindent \textbf{(D) Budgets:} 
For novel intent class discovery, a minimum number of human annotation is necessary. For NCD to work, at least 4 shot, 6-shot and 7-shot annotations are required for HWU, BANKING and CLINC datasets respectively. 

\begin{table*}[!bht]
\centering
\begin{adjustbox}{width=0.98\linewidth}
\begin{tabular}{|c|cccccccccccccccccc|}
\hline
\multirow{3}{*}{\textbf{Dataset}} & \multicolumn{8}{c|}{\textbf{Unsupervised Clustering Algorithms}}                                                                                                                                                                                                                                                    & \multicolumn{10}{c|}{\textbf{Semi-Supervised Clustering Algorithms}}                                                                                                                                                                                                                                                                                    \\ \cline{2-19} 
                                  & \multicolumn{2}{c|}{\textbf{KM}}                                        & \multicolumn{2}{c|}{\textbf{AG}}                                     & \multicolumn{2}{c|}{\textbf{DCN}}                                  & \multicolumn{2}{c|}{\textbf{DEC}}                                       & \multicolumn{2}{c|}{\textbf{DAL}}                                  & \multicolumn{2}{c|}{\textbf{DTC}}                                  & \multicolumn{2}{c|}{\textbf{KCL}}                                  & \multicolumn{2}{c|}{\textbf{MCL}}                                  & \multicolumn{2}{c|}{\textbf{CDAC+}}            \\ \cline{2-19} 
                                  & \multicolumn{1}{c|}{\textbf{A}}    & \multicolumn{1}{c|}{\textbf{F1}}   & \multicolumn{1}{c|}{\textbf{A}} & \multicolumn{1}{c|}{\textbf{F1}}   & \multicolumn{1}{c|}{\textbf{A}} & \multicolumn{1}{c|}{\textbf{F1}} & \multicolumn{1}{c|}{\textbf{A}}    & \multicolumn{1}{c|}{\textbf{F1}}   & \multicolumn{1}{c|}{\textbf{A}} & \multicolumn{1}{c|}{\textbf{F1}} & \multicolumn{1}{c|}{\textbf{A}} & \multicolumn{1}{c|}{\textbf{F1}} & \multicolumn{1}{c|}{\textbf{A}} & \multicolumn{1}{c|}{\textbf{F1}} & \multicolumn{1}{c|}{\textbf{A}} & \multicolumn{1}{c|}{\textbf{F1}} & \multicolumn{1}{c|}{\textbf{A}} & \textbf{F1} \\ \hline
\textbf{SNIPS}                    & \multicolumn{1}{c|}{\textbf{95.1}}              & \multicolumn{1}{c|}{\textbf{94.9}}              & \multicolumn{1}{c|}{92.7}           & \multicolumn{1}{c|}{92.9}              & \multicolumn{1}{c|}{89.2}           & \multicolumn{1}{c|}{88.7}            & \multicolumn{1}{c|}{89.6}              & \multicolumn{1}{c|}{88.2}              & \multicolumn{1}{c|}{92.2}           & \multicolumn{1}{c|}{92.2}            & \multicolumn{1}{c|}{87.6}           & \multicolumn{1}{c|}{87.2}            & \multicolumn{1}{c|}{73.3}           & \multicolumn{1}{c|}{70.4}            & \multicolumn{1}{c|}{78.2}           & \multicolumn{1}{c|}{74.1}            & \multicolumn{1}{c|}{80.4}           &     \multicolumn{1}{c|}{79.2}        \\ \hline
\textbf{ATIS}                     & \multicolumn{1}{c|}{\textbf{88.6}}              & \multicolumn{1}{c|}{\textbf{87.0}}              & \multicolumn{1}{c|}{85.8}           & \multicolumn{1}{c|}{86.4}              & \multicolumn{1}{c|}{77.7}           & \multicolumn{1}{c|}{79.78}            & \multicolumn{1}{c|}{83.1}              & \multicolumn{1}{c|}{85.42}              & \multicolumn{1}{c|}{86.9}           & \multicolumn{1}{c|}{87.0}            & \multicolumn{1}{c|}{84.3}           & \multicolumn{1}{c|}{85.9}            & \multicolumn{1}{c|}{76.7}           & \multicolumn{1}{c|}{80.8}            & \multicolumn{1}{c|}{80.4}           & \multicolumn{1}{c|}{83.2}            & \multicolumn{1}{c|}{77.9}           &        \multicolumn{1}{c|}{81.6}     \\\hline
\textbf{HWU}                      & \multicolumn{1}{c|}{83.8}          & \multicolumn{1}{c|}{84.2}          & \multicolumn{1}{c|}{83.2}       & \multicolumn{1}{c|}{83.3}          & \multicolumn{1}{c|}{84.1}       & \multicolumn{1}{c|}{83.6}        & \multicolumn{1}{c|}{\textbf{84.7}} & \multicolumn{1}{c|}{84.4} & \multicolumn{1}{c|}{83.7}       & \multicolumn{1}{c|}{82.6}        & \multicolumn{1}{c|}{83.6}       & \multicolumn{1}{c|}{\textbf{85.2}}        & \multicolumn{1}{c|}{73.3}       & \multicolumn{1}{c|}{74.1}        & \multicolumn{1}{c|}{78.1}       & \multicolumn{1}{c|}{74.8}        & \multicolumn{1}{c|}{83.1}       & 81.1        \\ \hline
\textbf{BANKING}                  & \multicolumn{1}{c|}{\textbf{84.7}} & \multicolumn{1}{c|}{\textbf{84.4}} & \multicolumn{1}{c|}{84.2}       & \multicolumn{1}{c|}{84.1} & \multicolumn{1}{c|}{80.1}       & \multicolumn{1}{c|}{83.2}        & \multicolumn{1}{c|}{80.1}          & \multicolumn{1}{c|}{80.5}          & \multicolumn{1}{c|}{80.5}       & \multicolumn{1}{c|}{81.1}        & \multicolumn{1}{c|}{79.9}       & \multicolumn{1}{c|}{78.2}        & \multicolumn{1}{c|}{71.8}       & \multicolumn{1}{c|}{72.4}        & \multicolumn{1}{c|}{74.2}       & \multicolumn{1}{c|}{73.1}        & \multicolumn{1}{c|}{83.4}       & 82.6        \\ \hline
\textbf{CLINC}                    & \multicolumn{1}{c|}{\textbf{94.9}} & \multicolumn{1}{c|}{\textbf{95.2}} & \multicolumn{1}{c|}{93.9}       & \multicolumn{1}{c|}{94.8}          & \multicolumn{1}{c|}{93.4}       & \multicolumn{1}{c|}{94.2}        & \multicolumn{1}{c|}{93.4}          & \multicolumn{1}{c|}{94.9}          & \multicolumn{1}{c|}{93.9}       & \multicolumn{1}{c|}{92.6}        & \multicolumn{1}{c|}{93.9}       & \multicolumn{1}{c|}{93.2}        & \multicolumn{1}{c|}{83.4}       & \multicolumn{1}{c|}{84.5}        & \multicolumn{1}{c|}{81.0}       & \multicolumn{1}{c|}{82.3}        & \multicolumn{1}{c|}{92.1}       & 92.5        \\ \hline
\end{tabular}
\end{adjustbox}
\caption{Accuracy (A) and F1-Score (F1) in (\%) for various Open Intent Discovery Based Clustering Algorithms across all datasets. The best results for each dataset in \textbf{Bold}.}
\label{tab:clustering_comp}
\end{table*}



\if{0}

All these clustering methods however cannot determine the intent classes of these clusters, we as such continue with the rest of our pipeline as discussed before, to determine the same. We use `bert-base-uncased' to extract representations for all the methods except DEC and DCN where we use Stacked Auto-Encoders. For NCD, we  consider $x$ as 2. For CBQA, we identify good cluster and bad cluster based on $p$ point selection from each cluster. If all the $p$ points from a cluster belong to same class then it is a good cluster otherwise bad cluster. The idea of good and bad cluster is helpful in prudent data selection for annotation. Good clusters are organised properly while bad clusters are cluttered. So less number of point selection is required from good clusters but more number of points are needed to be checked from bad clusters to optimize the labeling cost. For bad cluster, we select $q$ more points for annotation. In CBQA, we set $p$ as 3 and $q$ as 2. \pg{But later on, a different value has been given.} \textcolor{green}{[done]} We perform various ablation study on these values in the later sections. The results of all these methods are shown in Table \ref{tab:clustering_comp}. KM performs consistently well on all the datasets as can be observed in Table \ref{tab:clustering_comp}.

\fi

\noindent 

\begin{table*}[!b]
\centering
\begin{adjustbox}{width=\linewidth}
\begin{tabular}{|c|c|c|cccccccccc|}
\hline
\multirow{2}{*}{\textbf{Method}} & 
\multirow{2}{*}{\textbf{Silver Strategy}} & \multirow{2}{*}{\textbf{Gold Strategy}}              
& \multicolumn{2}{c|}{\textbf{SNIPS}}                                     & \multicolumn{2}{c|}{\textbf{ATIS}}                                      &
                                 \multicolumn{2}{c|}{\textbf{HWU}} 
                                                            &
                                 \multicolumn{2}{c|}{\textbf{BANKING}}                                       & 
                                 \multicolumn{2}{c|}{\textbf{CLINC}}                   \\ \cline{4-13} 
                                 &                                                                         &                                                                       & \multicolumn{1}{c|}{\textbf{A}}     & \multicolumn{1}{c|}{\textbf{F1}}    & \multicolumn{1}{c|}{\textbf{A}}     & \multicolumn{1}{c|}{\textbf{F1}}    & 
                                 \multicolumn{1}{c|}{\textbf{A}}     & \multicolumn{1}{c|}{\textbf{F1}}    & 
                                 \multicolumn{1}{c|}{\textbf{A}}     & \multicolumn{1}{c|}{\textbf{F1}}    & 
                                 \multicolumn{1}{c|}{\textbf{A}}     & \textbf{F1}    \\ \hline
MNID-1                           & 
Good Clusters$\dagger$ 
& \xmark                        &
\multicolumn{1}{c|}{94.4}          & \multicolumn{1}{c|}{93.2}          &  \multicolumn{1}{c|}{87.2}          & \multicolumn{1}{c|}{85.4}           &
\multicolumn{1}{c|}{78.5}           & \multicolumn{1}{c|}{78.8}       & \multicolumn{1}{c|}{77.5}           & \multicolumn{1}{c|}{78.4}           & \multicolumn{1}{c|}{89.2}           & \multicolumn{1}{c|}{89.8}          \\ \hline
MNID-2                           &
Good Clusters$\dagger$  & Any Point from Bad Clusters 
   & \multicolumn{1}{c|}{94.4}          & \multicolumn{1}{c|}{93.2}          &  \multicolumn{1}{c|}{87.2}          & \multicolumn{1}{c|}{85.4}                      & 
\multicolumn{1}{c|}{80.9}           & \multicolumn{1}{c|}{80.9}   & \multicolumn{1}{c|}{79.3}           & \multicolumn{1}{c|}{80.0}           & \multicolumn{1}{c|}{90.8}           & \multicolumn{1}{c|}{90.7}                  \\ \hline
MNID-3                           & \xmark                & 
Low-Conf from Any Cluster  
    & \multicolumn{1}{c|}{94.7}              & \multicolumn{1}{c|}{94.0}              & \multicolumn{1}{c|}{87.9}              &  \multicolumn{1}{c|}{86.1}          & \multicolumn{1}{c|}{81.2}           & \multicolumn{1}{c|}{81.1}          & \multicolumn{1}{c|}{81.7}           & \multicolumn{1}{c|}{81.1}           & \multicolumn{1}{c|}{91.3}           & \multicolumn{1}{c|}{91.0}          \\ \hline

MNID-4                           &
High-Conf from Good Clusters
&  \xmark                                       &
\multicolumn{1}{c|}{94.8} & \multicolumn{1}{c|}{93.9} & 
\multicolumn{1}{c|}{87.7} & \multicolumn{1}{c|}{86.1} &
\multicolumn{1}{c|}{81.5}           & \multicolumn{1}{c|}{81.4}      & \multicolumn{1}{c|}{82.2}           & \multicolumn{1}{c|}{81.8}           & \multicolumn{1}{c|}{91.8}           & \multicolumn{1}{c|}{91.9}      \\ \hline

MNID-5                           &
High-Conf from Good Clusters
& Low-Conf from Any Cluster                                        &
\multicolumn{1}{c|}{\textbf{95.1}} & \multicolumn{1}{c|}{\textbf{94.9}} & 
\multicolumn{1}{c|}{\textbf{88.6}} & \multicolumn{1}{c|}{\textbf{87.0}} &
\multicolumn{1}{c|}{82.9}           & \multicolumn{1}{c|}{82.2}      & \multicolumn{1}{c|}{82.7}           & \multicolumn{1}{c|}{81.8}           & \multicolumn{1}{c|}{92.1}           & \multicolumn{1}{c|}{93.5}      \\ \hline

MNID-6                           & Good Clusters$\dagger$                                                            & 
Low-Conf from Bad Clusters 
&
\multicolumn{1}{c|}{94.4} & \multicolumn{1}{c|}{93.2} & 
\multicolumn{1}{c|}{87.2} & \multicolumn{1}{c|}{85.4}             & 
\multicolumn{1}{c|}{83.1}           & \multicolumn{1}{c|}{82.8}        & \multicolumn{1}{c|}{83.9}          & \multicolumn{1}{c|}{83.1}          & \multicolumn{1}{c|}{93.9}          & \multicolumn{1}{c|}{93.7}          \\ \hline
MNID-7                           & \xmark                                                            & 
Low-Conf from Bad Clusters*
&
\multicolumn{1}{c|}{94.7} & \multicolumn{1}{c|}{94.0} & 
\multicolumn{1}{c|}{87.9} & \multicolumn{1}{c|}{86.1} & 
\multicolumn{1}{c|}{81.9}           & \multicolumn{1}{c|}{81.6}          & \multicolumn{1}{c|}{83.0}          & \multicolumn{1}{c|}{82.4}          & \multicolumn{1}{c|}{92.8}          & \multicolumn{1}{c|}{92.7}                  \\ \hline
MNID-8                           & 
Good Clusters$\dagger$ & Low-Conf from Bad Clusters* 

&
\multicolumn{1}{c|}{94.9} & \multicolumn{1}{c|}{94.4} & 
\multicolumn{1}{c|}{88.2} & \multicolumn{1}{c|}{86.4} & 
\multicolumn{1}{c|}{83.1}           & \multicolumn{1}{c|}{82.8} & \multicolumn{1}{c|}
{83.9} & \multicolumn{1}{c|}{83.1} & \multicolumn{1}{c|}{93.9} & \multicolumn{1}{c|}{93.7}                    \\ \hline
MNID-9                           & 
High-Conf from Good Clusters & Low-Conf from Bad Clusters* &
\multicolumn{1}{c|}{\textbf{95.1}} & \multicolumn{1}{c|}{\textbf{94.9}} & 
\multicolumn{1}{c|}{\textbf{88.6}} & \multicolumn{1}{c|}{\textbf{87.0}} & 
\multicolumn{1}{c|}{\textbf{83.8}}           & \multicolumn{1}{c|}{\textbf{83.2}} & \multicolumn{1}{c|}
{\textbf{84.7}} & \multicolumn{1}{c|}{\textbf{84.4}} & \multicolumn{1}{c|}{\textbf{94.9}} & \multicolumn{1}{c|}{\textbf{95.2}}                    \\ \hline
\end{tabular}
\end{adjustbox}
\caption{Accuracy (A) and F1-score (F1) in (\%) across all datasets for different variations of silver and gold strategy of MNID. [* - If no bad cluster exists then the strategy is applied on good clusters (SNIPS, ATIS have no bad cluster). Detailed  in line 6-11 of Algorithm \ref{algo4:annotation}. $\dagger$ We use $\mathcal{TH}$ = 0 in Algorithm \ref{algo4:annotation}.  \xmark : denotes we are not using this. \textbf{Bold} notifies the best for each dataset.]}
\label{tab:mnid-variations-table}
\end{table*}

\section*{Different Variations of 
MNID }

MNID consists of three steps (a). novel class detection (NCD), (b). cluster quality based annotation (CQBA) and (c). post-processing annotation strategy (PPAS). In each of these steps, certain parameters can be varied. We systematically discuss the impact of these parameters on MNID performance.

\vspace{-1mm}
\subsection*{Variations at NCD}

\noindent \textbf{(a) Performance of Clustering Algorithms:} 
We explore different unsupervised and semi-supervised clustering algorithms in our \textbf{MNID} framework. Overall accuracy and F1-Score for open intent discovery by different approaches 
are shown in Table \ref{tab:clustering_comp}. From Table \ref{tab:clustering_comp}, it is seen that unsupervised approaches perform better than semi-supervised models 
{The semi-supervised techniques get biased by the initial seed and fail to discover diverse clusters needed to detect all the new intent classes. }   K-Means (KM) performs the best across all datasets in terms of accuracy and F1 score except for HWU dataset where DEC and DTC (F1 only) outperforms it. 
This is most probably due to its robustness and absence of any outlier in the dataset. So we use K-Means as the clustering algorithm for MNID. 

\begin{figure}[h]
\centering
        \begin{subfigure}[b]{0.25\textwidth}
                \centering \includegraphics[width=0.99\linewidth]{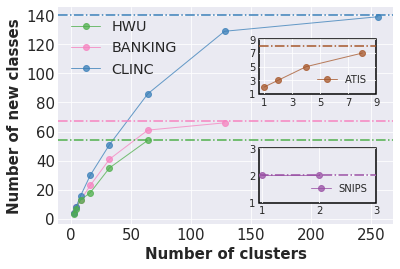}
                \caption{Class discovery with \\number of clusters}
                \label{fig:class_dis_1}
        \end{subfigure}%
        \begin{subfigure}[b]{0.25\textwidth}
                \centering
    \includegraphics[width=0.96\linewidth]{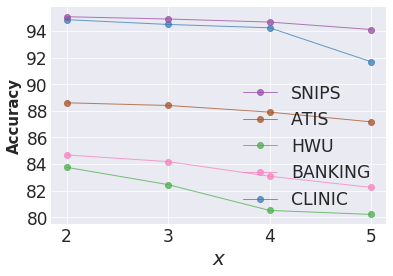}
                \caption{Accuracy vs points annotated ($x$) for clustering}
                
                \label{fig:x_1}
        \end{subfigure}
        \caption{Variations of NCD}
        \label{fig:NCD}
\end{figure}

\noindent \textbf{(b) Class Discovery with number of clusters:} From Fig \ref{fig:class_dis_1}, we observe an increasing trend in the number of classes discovered with increasing number of clusters which show that classes get evenly distributed across clusters as the number of clusters increases. The rate at which new classes are discovered is linear with the new clusters until significant classes are detected. The horizontal lines represent the gold number of new intents.

\begin{table}[!h]
\vspace{-2mm}
\centering
\begin{adjustbox}{width=0.9\linewidth}
\begin{tabular}{|c|c|c|c|c|c|c|c|}
\hline
p, q  & 2, 1 & 2, 2 & 2, 3 & 3, 0 & 3, 1  & \textbf{3, 2 }  & 4, 1     \\\hline
HWU & 82.1 & 81.2 & 82.5 & 81.8& 83.7 & \textbf{83.8} & 83.2 \\\hline
BANKING & 84.1 & 83.0 & 84.2 & 82.9 & 84.0 & \textbf{84.7} & 84.1 \\\hline
CLINC & 94.8 & 93.9 & 94.2 & 92.7& 93.0 & \textbf{94.9} &94.2 \\\hline
\end{tabular}
\end{adjustbox}
\caption{Accuracy (\%) based on point \\selections from Good and Bad clusters}
\label{tab:p_q_pointwise_accuracy1}
\end{table}

\noindent \textbf{(c) Effect of number of points ($x$)  used in clustering}: Fig \ref{fig:x_1} shows that the accuracy on all datasets drops as we increase the number of points used for new class discovery in clustering beyond $x=2$. This is because most of the budget gets exhausted while clustering and we have a very small budget to annotate low-confidence points in the next steps. Note that at least two points from a cluster need to be annotated for new class discovery.

\subsection*{Variations at CQBA}

\noindent \textbf{(a) Effect of number of points selected from Good and Bad clusters}: We experiment with different values of point selection ($p,q$) for the module CQBA (\ref{algo3:quality_cluster}) and observe how accuracy changes for three larger datasets - HWU, BANKING and CLINC. We get the best accuracy for ($p, q$) = (3, 2) i.e 3($p$) points from good cluster and 5($p+q$) points from bad cluster as shown in Table \ref{tab:p_q_pointwise_accuracy1}. Since, we perform gold annotation strategy on the bad clusters, a higher number of point selection is required to identify classes. 

\noindent \textbf{(b) Distribution of good and bad clusters}: For CLINC, BANKING and HWU we obtain 256, 128 and 64 clusters respectively by NCD. The percentage of good clusters obtained for CLINC, BANKING and HWU are  70.70\% (181 out of 256), 
46.88\% (60 out of 128) 
56.25 \% (36 out of 64), respectively. 
For BANKING, since the entire dataset is from a single domain with multiple intents being similar among themselves, we obtain more bad clusters than the good clusters. For SNIPS and ATIS, however, all the clusters are good clusters.

\subsection*{Variations at PPAS}

\noindent \textbf{(a) Different Variations of Gold and Silver Strategies:} The results for different variations of MNID methods (based on Silver and Gold Strategy applications) for all the datasets are provided in Table \ref{tab:mnid-variations-table}. 
We observe that the best result is obtained on MNID-9, i.e., choosing high confidence points from the good clusters for silver strategy and low confidence points from the bad clusters (if detected or else from the good clusters) only for gold strategy.
This strategy ensures that during silver annotation we choose points with high fidelity and side by side for gold annotation choose points with high uncertainty, both of which help in developing a highly accurate classifier. 
Silver strategy on high confidence points from good cluster (7 vs 9) and gold strategy on low confidence points from bad cluster (4 vs 9) alone enhances $\sim$1- 3\% accuracy and F1 for the three large datasets.
MNID-9 corresponds to our proposed approach, MNID.

\begin{table}[t]
\centering
\begin{adjustbox}{width=\linewidth}

\begin{tabular}{|c|c|c|c|c|c|}

\hline
\textbf{Silver Strategy on} & \textbf{SNIPS}
& \textbf{ATIS}  & \textbf{HWU} &  \textbf{BANKING} & \textbf{CLINC} \\ \hline
All Clusters (\%)     & 96.2 
& 95.9  & 82.5               & 84.3             & 85.2         \\ \hline
Good Clusters (\%)  & 96.2 & 95.9 &          93.6   & 95.4 & 97.2    
   \\ \hline
High-Conf from Good Clusters (\%)      & \textbf{97.8}    & \textbf{96.2} &\textbf{95.6}             & \textbf{97.2}  & \textbf{98.1} 
\\\hline\hline
Average Points per Class & 10.1 & 8.2 & 17.0 & 22.8 &15.9 \\\hline
\end{tabular}
\end{adjustbox}
\caption{Accuracy (in \%) and usage of average \\number of datapoints per class in silver strategy}
\label{tab:silver}
\end{table}

\noindent \textbf{(b) Silver Strategy Analysis:} We inspect silver strategy based on cosine similarity, confidence score and strategy accuracy. 

\begin{figure}[!t]
\vspace{-4mm}
\centering
\captionsetup{justification=centering}
        \begin{subfigure}[b]{0.25\textwidth}
                \centering \includegraphics[width=.95\linewidth]{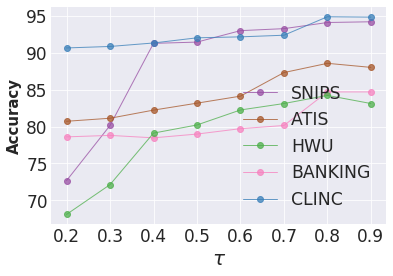}
                \caption{Cosine similarity \\threshold ($\tau$)}
                \label{fig:cos}
        \end{subfigure}%
        \begin{subfigure}[b]{0.25\textwidth}
                \centering
                \includegraphics[width=.95\linewidth]{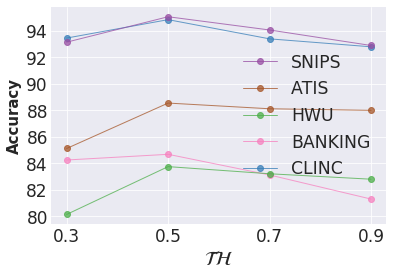}
                \caption{Confidence score \\threshold ($\mathcal{TH}$)}
                \label{fig:conf}
        \end{subfigure}%
        \caption{Variations of PPAS }\label{fig:variations}
\end{figure}

\textbf{(i) Effect of Cosine Similarity and Confidence Score (CS)}: 
We study the effect of cosine similarity of silver strategy for MNID. From Fig. \ref{fig:cos}, we observe that the best results are always obtained using a higher threshold of 0.8 cosine similarity. In case of BANKING, HWU and ATIS accuracy drops at 0.9 whereas for other datasets it remains almost identical. 
Fig \ref{fig:conf} shows how accuracy varies for different confidence scores. We observe that for all the datasets  
the best results are obtained at a threshold of 0.5. This is because a lower threshold allows more diverse datapoints to be selected using cosine similarity and this in turn improves the model performance. 
In both the cases if the cosine similarity or the threshold is increased beyond the optimal point, that results in selection of too less  datapoints which is not enough for the classifier to do a meaningful learning. Hence accuracy drops. So we choose the parameters - cosine similarity = 0.8 and $\tau$ = 0.5 - while choosing {\em high confidence} point to be annotated by silver strategy. 

 \textbf{(ii) Strategy Accuracy}: The accuracy of data point selection by silver strategy for different MNID variations is shown in Table \ref{tab:silver}. 
We see the strategy of choosing high-confidence points from good clusters produce points with high fidelity. 
Table \ref{tab:silver} also shows the average number of points per class as selected by this strategy for various datasets. Here we see that enough number of silver points are annotated even after considering a very strict criterion. 
Note, count is the highest for BANKING because multiple intents are very similar to each other and hence more points qualify the cosine similarity threshold, $\tau$. 

.

 In Table \ref{tab:fb_results}, we show results on Facebook multilingual datasets, ATIS and SNIPS (averaged results from Table \ref{tab:snips_results}). In Table \ref{tab:snips_results}, we report performance of our algorithm on the different types on SNIPS data. We observe that in almost all the datasets, the \textbf{KM} labelling strategy outperforms others in terms of overall accuracy and Macro-F1-score. We also observe that \textbf{CL} performs significantly well on all the datasets and even out-performs \textbf{KM} in SNIPS Type 2 setting as in Table \ref{tab:snips_results}. Thus, \textbf{CL} can be a decent alternative to \textbf{KM} and may significantly reduce the overhead of deciding the number of clusters. The MNID Random annotation (Rand Annot) baseline fails to perform well on the datasets (Table \ref{tab:snips_results}) since the random selection of samples leads to a uniform distribution of all the known and unknown classes and a lower number of OOD samples are added back to training.

\noteng{A detailed discussion is needed}





Our method (MNID) consistently outperforms the other baseline by $\sim$ 10\% in terms of accuracy and $\sim$ 4-8\% in terms of macro F1. Also, it can be observed from Table \ref{tab:snips_results}, that all the baselines perform well for Type 3 setting in the SNIPS dataset. The reason behind this would be the high level of dissimilarity between the unknown intents with the known intents which makes it easier to detect these classes. Despite of the fact that the competing baselines have the added advantage of not having to differentiate between the novel classes, we were still able to achieve higher results on all the datasets. 

\label{sec:ablation}

\if{0}
\noindent \textbf{(g) Effect of changing size of unlabelled data}: We study the effect of changing unlabelled data on three larger datasets viz., BANKING, CLINC and HWU. We split the dataset into: 2x, 5x and Full setting where x is the budget. These data points are chosen randomly from the pool of unlabelled data available. We observe that the performance of our method increases with the size of the unlabelled data as can be seen in Fig \ref{fig:unlab}. The dotted lines represent the accuracy of this method in `Gold10' setting. 

\noteng{How are you choosing 2x, 5x etc. What is the percentage of classes you are being able to discover} \textcolor{green}{[done]}\\

\fi


\vspace{-3mm}
\section{Conclusion}
\label{sec:conclusion}
We have developed MNID (\textbf{M}ultiple \textbf{N}ovel \textbf{I}ntent \textbf{D}etection), an end-to-end framework 
 to identify multiple novel intents 
within a fixed annotation cost. 
The algorithm intelligently uses the concept of clusters to first discover the classes and then  estimate the nature in which datapoints of a class is distributed, that is, whether the datapoints of a class congregate strongly within themselves and separate from other classes or are entangled with datapoints of other classes. In the two types of situations, we propose two different strategies, silver strategy to take advantage of the clusters so that we can annotate many points without any extra human cost and gold strategy to annotate highly uncertain points. 
This two-pronged approach helps us to annotate 
highly precise points automatically while annotating the most uncertain (with respect to the class it belongs) points using human assistance. 
We have done a very rigorous analysis/experimentation to establish the core idea of our algorithm. 
We observe that the accuracy of classifiers when fed with the dataset created by MNID can beat the standard best few-shot setting where it is assumed that `$\kappa$' instances of each class are provided and annotated by human whereas in our case we have to first discover the classes and then have to find the instances of each class. 

One limitation of MNID is that it is not able to detect intents where classes are very similar to each other. For example, the query ``Can you explain why my payment is still pending?'' in BANKING dataset is from the ``pending transfer'' category but our system detects as ``pending card payment’’ intent as both intents are quite similar. We shall try to address this issue in future.
We have presently worked on a setting where novel intents appear in one step, we would strive to extend this framework 
to explore the dynamics of periodically evolving intents. 

\bibliography{8custom}
\bibliographystyle{acl_natbib}
\newpage
\end{document}